\documentclass{article}

\usepackage{times}
\usepackage{graphicx} 
\usepackage{subfigure} 

\usepackage{natbib}

\usepackage{algorithm}
\usepackage{algorithmic,verbatim}

\usepackage{amsmath,amssymb}
\usepackage{alltt}

\usepackage{hyperref}

\newcommand{\ie}{\emph{i.e.}}
\newcommand{\eg}{\emph{e.g.}}

\newcommand{\diff}[2]{\frac{\partial #1}{\partial #2}}
\newcommand{\vct}[1]{\ensuremath{\boldsymbol{#1}}} 
\newcommand{\mat}[1]{\ensuremath{\mathbf{#1}}}
\newcommand{\set}[1]{\ensuremath{\mathcal{#1}}}

\newcommand{\T}{\ensuremath{\top}}

\usepackage[accepted]{icml2015}

\icmltitlerunning{Is Feature Selection Secure against Training Data Poisoning?}

\begin{document} 

\twocolumn[
\icmltitle{Is Feature Selection Secure against Training Data Poisoning?}

\icmlauthor{Huang Xiao}{xiaohu@in.tum.de}
\icmladdress{Department of Computer Science, Technische Universit\"at M\"unchen, Boltzmannstr.3, 85748 Garching, Germany}
\icmlauthor{Battista Biggio}{battista.biggio@diee.unica.it}
\icmladdress{Department of Electrical and Electronic Engineering, University of Cagliari, Piazza d'Armi, 09123 Cagliari, Italy}
\icmlauthor{Gavin Brown}{gavin.brown@manchester.ac.uk}
\icmladdress{School of Computer Science, University of Manchester, Oxford Road, M13 9PL, UK}
\icmlauthor{Giorgio Fumera}{fumera@diee.unica.it}
\icmladdress{Department of Electrical and Electronic Engineering, University of Cagliari, Piazza d'Armi, 09123 Cagliari, Italy}
\icmlauthor{Claudia Eckert}{claudia.eckert@in.tum.de}
\icmladdress{Department of Computer Science, Technische Universit\"at M\"unchen, Boltzmannstr.3, 85748 Garching, Germany}
\icmlauthor{Fabio Roli}{roli@diee.unica.it}
\icmladdress{Department of Electrical and Electronic Engineering, University of Cagliari, Piazza d'Armi, 09123 Cagliari, Italy}

\icmlkeywords{adversarial machine learning, feature selection, poisoning attacks}

\vskip 0.3in
]

\begin{abstract} 
Learning in adversarial settings is becoming an important task for application domains where attackers may inject malicious data into the training set to subvert normal operation of data-driven technologies. Feature selection has been widely used in machine learning for security applications to improve generalization and computational efficiency, although it is not clear whether its use may be beneficial or even counterproductive when training data are poisoned by intelligent attackers.
In this work, we shed light on this issue by providing a framework to investigate the robustness of popular feature selection methods, including LASSO, ridge regression and the elastic net. Our results on malware detection show that feature selection methods can be significantly compromised under attack (we can reduce LASSO to almost random choices of feature sets by careful insertion of less than 5\% poisoned training samples), highlighting the need for specific countermeasures.
\end{abstract} 

\section{Introduction}

With the advent of the modern Internet, the number of interconnected users and devices, along with the available number of services, has tremendously increased. This has not only simplified our lives, through accessibility and ease-of-use of novel services (\eg, think to the use of maps and geolocation on smartphones), but it has also provided great opportunities for attackers to perform novel and profitable malicious activities. 
To cope with this phenomenon, machine learning has been adopted in security-sensitive settings like spam and malware detection, web-page ranking and network protocol verification~\cite{sahami98,mccallum,rubinstein09,barreno10,smutz12,bruckner12,biggio12-icml,biggio13-aisec,biggio14-tkde}.
In these applications, the challenge is that of inferring actionable knowledge from a large, usually high-dimensional data collection, to correctly prevent malware (\ie, malicious software) infections or other threats. For instance, detection of malware in PDF files relies on the analysis of the PDF logical structure, which consists of a large set of different kinds of objects and metadata, yielding a high-dimensional data representation~\cite{smutz12,maiorca12-mldm,maiorca13-asiaccs,srndic13-ndss}. Similarly, text classifiers for spam filtering rely on the construction of a large dictionary to identify words that are mostly discriminant of spam and legitimate emails~\cite{sahami98,mccallum,graham02,robinson03}.

Due to the large number of available features, learning in these tasks is particularly challenging. Feature selection is thus a crucial step for  reducing the impact of the curse of dimensionality on classifier's generalization, and for learning  efficient models providing easier-to-interpret decisions.

For the same reasons behind the growing sophistication and variability of modern attacks, it is reasonable to expect that, being increasingly adopted in these tasks, machine learning techniques will be soon targeted by specific attacks, crafted by skilled attackers. In the last years, relevant work in the area of adversarial machine learning has addressed this issue, and proposed some pioneering methods for secure learning against particular kinds of attacks~\cite{barreno06-asiaccs,huang11,biggio14-tkde,biggio12-icml,biggio13-ecml,bruckner12,globerson06-icml}.

While the majority of work has focused on analyzing vulnerabilities of classification and clustering algorithms, only recent work has considered intrinsic vulnerabilities introduced by the use of feature selection methods.
In particular, it has been shown that classifier evasion can be facilitated if features are not selected according to an adversary-aware procedure that explicitly accounts for adversarial data manipulation at test time~\cite{bo14-nips,wang14-icdm,zhang15-tcyb}. Although these attacks do not directly target feature selection, but rather the resulting classification system, they highlight the need for adversarial feature selection procedures. Attacks that more explicitly target feature selection fall into the category of \emph{poisoning attacks}. Under this setting, the attacker has access to the training data, and  contaminates it to subvert or control the selection of the reduced feature set. 

As advocated in a recent workshop~\cite{joseph13-dagstuhl}, poisoning attacks are an emerging security threat for data-driven technologies, and could become the most relevant one in the coming years, especially in the so-called big-data scenario dominated by data-driven technologies.
From a practical perspective, poisoning attacks are already a pertinent scenario in several applications. For instance, in collaborative spam filtering, classifiers are retrained on emails labeled by end users. Attackers owning an authorized email account protected by the same anti-spam filter may thus arbitrarily manipulate emails in their inbox, \ie, part of the training data used to update the classifier.
Some systems may even ask directly to users to validate their decisions on some submitted samples, and use their feedback to update the classifier (see, \eg, PDFRate, an online tool for detecting PDF malware designed by~\citealp{smutz12}).
Furthermore, in several cases obtaining accurate labels, or validating the available  ground truth may be expensive and time-consuming; \eg, if malware samples are collected from the Internet, by means of honeypots, \ie, machines that purposely expose known vulnerabilities to be infected by malware~\cite{spitzner02}, or other online services, like VirusTotal,\footnote{\url{http://virustotal.com}} labeling errors are possible.

Work by \citet{rubinstein09} and \citet{nelson08} has shown the potential of poisoning attacks against PCA-based malicious traffic detectors and spam filters, and proposed robust techniques to mitigate their impact.
More recently, \citet{mei15-aistats} have demonstrated how to poison latent Dirichlet allocation to drive its selection of relevant topics towards the attacker's choice. 

In this work, we propose a framework to categorize and provide a better  understanding of the different attacks that may target feature selection algorithms, building on previously-proposed attack models for the security evaluation of supervised and unsupervised learning algorithms~\cite{biggio14-tkde,biggio13-aisec,biggio12-icml,huang11,barreno06-asiaccs} (Sect.~\ref{sect:framework}).
We then exploit this framework to formalize poisoning attack strategies against popular embedded feature selection methods, including the so-called \emph{least absolute shrinkage and selection operator} (LASSO)~\cite{tibshirani96-lasso}, \emph{ridge} regression~\cite{hoerl70-ridge}, and the \emph{elastic net}~\cite{zou05-elasticnet} (Sect.~\ref{sect:poisoning}). 
We report experiments on PDF malware detection, assessing how poisoning affects both feature selection and classification error (Sect.~\ref{sect:exp}).
We conclude the paper by discussing our findings and contributions (Sect.~\ref{sect:discussion}), and sketching promising future research directions (Sect.~\ref{sect:conclusion}).

\section{Feature Selection Under Attack} \label{sect:framework}

In this section, we present our framework for the security evaluation of feature selection algorithms. 
It builds on the framework proposed by \citet{biggio14-tkde,biggio13-aisec} to assess the security of classification and clustering algorithms, which in turn relies on a taxonomy of attacks against learning algorithms originally proposed by \citet{huang11,barreno06-asiaccs}. Following the framework of \citet{biggio14-tkde,biggio13-aisec}, we define ours in terms of assumptions on the attacker's goal, knowledge of the system, and capability of manipulating the input data.

\textbf{Notation.} In the following, we assume data is generated according to an underlying i.i.d.~process $p: \set X \mapsto \set Y$, for which we are only given a set $\set D = \{\vct x_{i}, y_{i} \}_{i=1}^{n}$ of $n$ samples, each consisting of a $d$-dimensional feature vector $\vct x_{i} = [ x_{i}^{1}, \ldots, x_{i}^{d} ]^{\T} \in \set X$, and a target variable $y_{i} \in \set Y$. 
Learning amounts to inferring the underlying process $p$ from $\set D$.
Feature selection can be exploited to facilitate this task by selecting a suitable, relevant feature subset from $\set D$, according to a given criterion. For instance, although in different forms, wrapper and embedded methods aim to minimize classification error, while information theoretic filters optimize different estimates of the information gain~\cite{brown12}.
To denote a given feature subset, we introduce a vector $\vct \pi \in \{0,1\}^{d}$, where each element denotes whether the corresponding feature has been selected (1) or not (0). Then, a feature selection algorithm can be represented in terms of a function $h(\set D)$ that selects a feature subset $\vct \pi$ by minimizing a given selection criterion $\set L(\set D, \vct \pi)$ (\eg, the classification error). 


\subsection{Attacker's Goal}
\label{sect:attacker-goal}

The attacker's goal is defined in terms of the desired \emph{security violation}, which can be an \textbf{integrity}, \textbf{availability}, or \textbf{privacy} violation, and of the \emph{attack specificity}, which can be \textbf{targeted} or \textbf{indiscriminate}~\citep{barreno06-asiaccs,huang11,biggio14-tkde,biggio13-aisec}.

\textbf{Integrity} is violated if malicious activities are performed without compromising normal system operation, \eg, attacks that evade classifier detection without affecting the classification of legitimate samples. 
In the case of feature selection, we thus regard integrity violations as attacks that only slightly modify the selected feature subset, aiming to facilitate subsequent evasion; \eg, an attacker may aim to avoid the selection of some specific words by an anti-spam filter, as they are frequently used in her spam emails.

\textbf{Availability} is violated if the functionality of the system is compromised, causing a denial of service. 
For classification and clustering, this respectively amounts to causing the largest possible classification error and to maximally altering the clustering process on the input data~\cite{huang11,biggio14-tkde,biggio13-aisec}.
Following the same rationale, the availability of a feature selection algorithm is compromised if the attack enforces selection of a feature subset which yields the largest generalization error.

\textbf{Privacy} is violated if the attacker is able to obtain information about the system's users by reverse-engineering the attacked system. In our case, this would require the attacker to reverse-engineer the feature selection process, and, getting to know the selected features, infer information about the training data and the system users.\footnote{Note that privacy attacks against machine learning are very speculative, and only considered here for completeness. Further, feature selection algorithms are typically unstable, making them very difficult to reverse-engineer. It would be thus of interest to understand whether feature selection exhibits some privacy guarantees, \eg, if it can be intrinsically differentially private.}

Finally, the attack specificity is \textbf{targeted}, if the attack affects the selection of a \emph{specific} feature subset, and \textbf{indiscriminate}, if the attack affects the selection of \emph{any} feature.

\subsection{Attacker's Knowledge}
\label{sect:attacker-knowledge}

The attacker can have different levels of knowledge of the system, according to specific assumptions on the points ($k.i$)-($k.iii$) described in the following.

\textbf{$(k.i)$ Knowledge of the training data $\set D$}: 
The attacker may have partial or full access to the training data $\set D$.
If no access is possible, she may collect a \emph{surrogate} dataset $\hat{\set D} = \{\hat{\vct x}_{i}, \hat y_{i}\}_{i=1}^{m}$, ideally drawn from the same underlying process $p$ from which $\set D$ was drawn, \ie, from the same source from which samples in $\set D$ were collected; \eg, honeypots for malware samples~\cite{spitzner02}.

\textbf{$(k.ii)$ Knowledge of the feature representation $\set X$}:
The attacker may partially or fully know how features are computed for each sample, before performing feature selection.

\textbf{$(k.iii)$ Knowledge of the feature selection algorithm $h(\set D)$}: 
The attacker may know that a specific feature selection algorithm is used, along with a specific selection criterion $\set L(\set D)$; \eg, the accuracy of a given classifier, if a wrapper method is used. In a very pessimistic setting, the attacker may even discover the selected feature subset.

\textbf{Perfect Knowledge (PK).} The worst-case scenario in which the attacker has full knowledge of the attacked system, is usually referred to as \emph{perfect knowledge} case \cite{biggio14-tkde,biggio13-aisec,biggio12-icml,kloft10,bruckner12,huang11,barreno06-asiaccs}, and it allows one to empirically evaluate an upper bound on the performance degradation that can be incurred by the system under attack. In our case, it amounts to completely knowing: $(k.i)$ the data, $(k.ii)$ the feature set, and $(k.iii)$ the feature selection algorithm.

\textbf{Limited Knowledge (LK).}  Attacks with \emph{limited knowledge} have also been often considered, to simulate more realistic settings~\cite{biggio14-tkde,biggio13-aisec,biggio13-ecml}. 
In this case, the attacker is assumed to have only partial knowledge of $(k.i)$ the data, \ie, she can collect a surrogate dataset $\hat{\set D}$, but knows $(k.ii)$ the feature set, and $(k.iii)$ the feature selection algorithm.
She can thus replicate the behavior of the attacked system using the surrogate data $\hat{\set D}$, to construct a set of attack samples. The effectiveness of these attacks is then assessed against the targeted system (trained on $\set D$).

\subsection{Attacker's Capability}
\label{sect:attacker-capability}

The attacker's capability defines how and to what extent the attacker can control the feature selection process.
As for supervised learning, the attacker can influence both training and test data, or only test data, respectively exercising a \textbf{causative} or \textbf{exploratory} influence (more commonly referred to as poisoning and evasion attacks)~\cite{barreno06-asiaccs,huang11,biggio14-tkde}.
Although modifying data at test time does not affect the feature selection process directly, it may nevertheless influence the security of the corresponding classifier against evasion attacks at test time, as also highlighted in recent work~\cite{bo14-nips,wang14-icdm}. Therefore, evasion should be considered as a plausible attack scenario even against feature selection algorithms. 

In poisoning attacks, the attacker is often assumed to control a small percentage of the training data $\set D$ by injecting a fraction of well-crafted attack samples. The ability to manipulate their feature values and labels depends on how labels are assigned to the training data; \eg, if malware is collected via \emph{honeypots}~\cite{spitzner02}, and labeled with some anti-virus software, the attacker has to construct poisoning samples under the constraint that they will be labeled as expected by the given anti-virus.

In evasion attacks, the attacker manipulates malicious data at test time to evade detection. Clearly, malicious samples have to be manipulated without affecting their malicious functionality, \eg, malware code has to be obfuscated without compromising the exploitation routine. In several cases, these constraints can be encoded in terms of distance measures between the original, non-manipulated attack samples and the manipulated ones \cite{dalvi04,lowd05,globerson06-icml,teo08,bruckner12,biggio13-ecml}.

\subsection{Attack Strategy}

Following the approach in~\citet{biggio13-aisec}, we define an \emph{optimal} attack strategy to reach the attacker's goal, under the constraints imposed by her knowledge of the system and capabilities of manipulating the input data. 
To this end, we characterize the attacker's knowledge in terms of a space $\Theta$ that encodes assumptions ($k.i$)-($k.iii$) on the knowledge of the data $\set D$, the feature space $\set X$, the feature selection algorithm $h$, and the selection criterion $\set L$. Accordingly, for PK and LK attacks, the attacker's knowledge can be respectively represented as $\vct \theta_{\rm PK}=(\set D,\set X,h, \set L)$ and $\vct \theta_{\rm LK}=(\hat{\set D},\set X,h, \set L)$. 
We characterize the attacker's capability by assuming that an initial set of samples $\set A$ is given, and that it is modified according to a space of possible modifications $\Phi(\set A)$.
Given the attacker's knowledge $\vct \theta\in\Theta$ and a set of manipulated attacks $\set A^{\prime} \in \Phi(\set A)$, the attacker's goal can be characterized in terms of an objective function $\set W (\set A^{\prime}, \vct \theta) \in \mathbb R$ which evaluates how effective the attacks $\set A^{\prime}$ are.
The optimal attack strategy can be thus given as: 
\begin{equation}
	\displaystyle
	\begin{array}{rl}
		\max_{\set A^{\prime}}&  \set W(\set A'; \vct \theta) \\
		\text{s.t.} &\set A'\in\Phi(\set A)\,.
	\end{array}
	\label{eq:optim}
\end{equation}

\subsection{Attack Scenarios}

Some relevant attack scenarios that can be formalized according to our framework are briefly sketched here, also mentioning related work. For the sake of space, we do not provide a thorough formulation of all these attacks. However, this can be obtained similarly to the formulation of poisoning attacks given in the next section.

\textbf{Evasion attacks.} Under this setting, the attacker's goal is to manipulate malicious data at test time to evade detection, as in the recent attacks envisioned by~\citet{bo14-nips,wang14-icdm} (\ie, an integrity, indiscriminate violation with exploratory influence). Although evasion does not affect feature selection directly, the aforementioned works have shown that selecting features without taking into account the adversarial presence may lead one to design much more vulnerable systems. Thus, evasion attacks should be considered as a potential scenario to explore vulnerabilities of classifiers learnt on reduced feature sets, and to properly design more secure, adversary-aware feature selection algorithms.

\textbf{Poisoning (integrity) attacks.} Here, we envisage another scenario in which the attacker tampers with the training data to enforce selection of a feature subset that will facilitate classifier evasion at test time (\ie, an integrity, targeted violation with causative influence). For instance, an attacker may craft poisoning samples to enforce selection of a given subset of features, whose values are easily changed with trivial manipulations to the malicious data at test time.

\textbf{Poisoning (availability) attacks.} Here the attacker aims to inject well-crafted poisoning points into the training data to maximally compromise the output of the feature selection algorithm, or directly of the learning algorithm~\cite{mei15-aistats,biggio12-icml,rubinstein09,nelson08} (\ie, an availability, indiscriminate violation with causative influence). This attack scenario is formalized in detail according to our framework in the next section, to target embedded feature selection algorithms.

\section{Poisoning Embedded Feature Selection}
\label{sect:poisoning}

We report now a detailed case study on poisoning (availability) attacks against embedded feature selection algorithms, including LASSO, ridge regression, and the elastic net.  These algorithms perform feature selection by learning a linear function $f(\vct x)=\vct w^{\T} \vct x + b$ that minimizes the trade-off between a loss function $\ell \left (y, f(\vct x) \right )$ computed on the training data $\set D$ and a regularization term $\Omega(\vct w)$. The selection criterion $\set L$ can be thus generally expressed as:
\begin{equation}
\label{eq:tykhonov}
\min_{\vct w, b} \; \set L =  \frac{1}{n}\sum_{i=1}^{n} \ell \left (y_{i}, f(\vct x_{i}) \right ) + \lambda \Omega(\vct w) \, ,
\end{equation}
where $\lambda$ is the trade-off parameter.\footnote{Note that this is equivalent to minimizing the loss subject to $\Omega(\vct w) \leq t$, for proper choices of $\lambda$ and $t$~\cite{tibshirani96-lasso}.}
The quadratic loss $\ell(y, f(\vct x)) = \frac{1}{2} \left ( f( \vct x)-y \right )^{2}$ is used by all the three considered algorithms.
As for the regularization term $\Omega(\vct w)$, ideally, one would like to consider the $\ell_{0}$-norm of $\vct w$ to exactly select a given number of features, which however makes the problem NP-hard~\cite{natajaran95}.
LASSO uses $\ell_{1}$ regularization, \ie, $\Omega(\vct w) = \| \vct w \|_{1}$, yielding the tighter convex relaxation to the ideal problem formulation.
Ridge regression uses $\ell_{2}$ regularization, $\Omega(\vct w) = \frac{1}{2}\| \vct w \|^{2}_{2}$. The elastic net is a hybrid approach between the aforementioned ones, as it exploits a convex combination of $\ell_{1}$ and $\ell_{2}$ regularization, \ie, $\Omega(\vct w) =  \rho \| \vct w \|_{1} + (1- \rho) \frac{1}{2} \| \vct w \|^{2}_{2}$, where $\rho \in (0,1)$.
Eventually, if one is given a maximum number of features $k < d$ to be selected, the ones corresponding to the first $k$ feature weights sorted in descending order of their absolute values can be thus retained.

In the considered setting, the \textbf{attacker's goal} is to maximally increase the classification error of these algorithms, by enforcing the selection of a wrong subset of features.
As for the \textbf{attacker's knowledge}, we consider both PK and LK as discussed in Sect.~\ref{sect:attacker-knowledge}. In the sequel, we consider LK attacks on the surrogate data $\hat{\set D}$, since for PK attacks we can simply set $\hat{\set D} =\set D$.
The \textbf{attacker's capability} amounts to injecting a maximum number of poisoning points into the training set.
To estimate the classification error, the attacker can evaluate the same criterion $\set L$ used by the embedded feature selection algorithm, on her available training set $\hat{\set D}$, excluding the attack points (as they will not be part of the test data).
The attack samples are thus kept outside from the empirical loss computation of the attacker, while they are clearly taken into account by the learning algorithm.
Assuming that a single attack point $\vct x_{c}$ is added by the attacker, the \textbf{attack strategy} can be thus formulated as:
\begin{equation}
\max_{\vct x_{c}} \; \set W = \frac{1}{m}\sum_{j=1}^{m} \ell \left (\hat y_{j}, f(\hat{\vct x}_{j}) \right ) + \lambda \Omega(\vct w)
\label{eq:attack-obj}
\end{equation} 
where it is worth remarking that $f$ is learnt by minimizing $\set L(\hat{\set D} \cup \{\vct x_{c}\})$ (Eq.~\ref{eq:tykhonov}), and thus depends on the attack point $\vct x_{c}$, as well as the corresponding $\vct w$ and b.
The attacker's objective $\set W$ (Eq.~\ref{eq:attack-obj}) can be thus optimized by iteratively modifying $\vct x_{c}$ with a (sub)gradient-ascent algorithm, in which, at each step, the solution $\vct w,b$ is updated by minimizing $\set L(\hat{\set D} \cup \{\vct x_{c}\})$, \ie, simulating the behavior of the feature selection algorithm on the poisoned data. Note that the parameters $\vct w,b$ estimated by the attacker are not generally the same ones estimated by the targeted algorithm. The latter will be indeed estimated by minimizing $\set L(\set D \cup \{\vct x_{c}\})$.

\paragraph{Gradient Computation.} By calculating the partial derivative of Eq.~\eqref{eq:attack-obj} with respect to $\vct x_{c}$, and substituting $\ell(y, f(\vct x))$ and $f$ with their expressions, one yields:
\begin{align}
\diff{\set W }{\vct x_{c}}=\frac{1}{m}\sum_{j=1}^{m} \left ( f(\hat{\vct x}_{j}) - \hat y_{j} \right ) \left (\hat{\vct x}_{j}^{\T}\diff{\vct w}{\vct x_{c}} + \diff{b}{\vct x_{c}} \right) + \lambda \vct r \diff{\vct w}{\vct x_{c}} \, ,
\label{eq:gradient-xc}
\end{align}
where, for notational convenience, we set $\vct r = \diff{\Omega}{\vct w}$.
Note that $\vct r = {\rm sub}(\vct w)$ for LASSO, $\vct r = \vct w$ for ridge regression, and  $\vct r = \rho \, {\rm sub}(\vct w) + (1-\rho) \vct w$ for the elastic net, being ${\rm sub}(\vct w)$ the subgradient of the $\ell_{1}$-norm, \ie, a vector whose $k^{\rm th}$ component equals $+1$ (-1) if $w^{k} > 0$ ($w^{k} < 0$), and any value in $[-1,+1]$ if $w^{k}=0$.
As the subgradient is not uniquely determined, a large set of possible ascent directions should be explored, dramatically increasing the computational complexity of the attack algorithm. 
Further, computing $\diff{\vct w}{\vct x_{c}}$ and $\diff{b}{\vct x_{c}}$ requires us to predict how the solution $\vct w,b$ changes while the attack point $\vct x_{c}$ is modified.

To overcome these issues, as in \citet{cauwenberghs00,biggio12-icml}, we assume that the Karush-Kuhn-Tucker (KKT) conditions under perturbation of the attack point $\vct x_{c}$ remain satisfied, \ie, we adjust the solution to remain at the optimum.
At optimality, the KKT conditions for Problem~\eqref{eq:tykhonov} with quadratic loss and linear $f$, are:
\begin{align}
\label{eq:KKT1}
& \diff{\set L}{\vct w}^{\T} = \frac{1}{n}\sum_{i=1}^{n} \left ( f(\hat{\vct x}_{i}) - \hat y_{i} \right) \hat{\vct x}_{i} + \lambda \vct r^{\T} = \vct 0\, ,\\
\label{eq:KKT2}
& \diff{\set L}{b} = \frac{1}{n}\sum_{i=1}^{n} \left ( f(\hat{\vct x}_{i}) - \hat y_{i} \right) =0\, ,
\end{align}
where we transposed the first equation to have a column vector, and keep the following derivation consistent.
If $\set L$ is convex but \emph{not differentiable} (\eg, when using $\ell_{1}$ regularization), one may express these conditions using subgradients. In this case, at optimality a necessary and sufficient condition is that at least one of the subgradients of the objective is null~\cite{Boyd-Vandenberghe-Convex-2004}.
In our case, at optimality, the subgradient is uniquely determined from Eq.~\eqref{eq:KKT1} as $\vct r = - \frac{1}{\lambda} \frac{1}{n}\sum_{i=1}^{n} \left ( f(\hat{\vct x}_{i}) - \hat y_{i} \right) \hat{\vct x}_{i}^{\T}$.
This allows us to drastically reduce the complexity of the attack algorithm, as we are not required to explore all possible subgradient ascent paths for Eq.~\eqref{eq:gradient-xc}, but just the one corresponding to the optimal solution. 

Let us assume that the optimality conditions given by Eqs.~\eqref{eq:KKT1}-\eqref{eq:KKT2} remain valid under the perturbation of $\vct x_{c}$. We can thus set their derivatives with respect to $\vct x_{c}$ to zero. After deriving and re-arranging in matrix form, one obtains:
\begin{eqnarray}
\label{eq:lin-sys}
\begin{bmatrix}
\mat \Sigma + \lambda \vct v & \vct \mu \\
\vct \mu^{\T} & 1
\end{bmatrix}
\begin{bmatrix}
 \diff{\vct w}{\vct x_{c}} \\
  \diff{b}{\vct x_{c}}
\end{bmatrix}
= -\frac{1}{n} \begin{bmatrix}
\mat M \\
\vct w^{\T} 
\end{bmatrix} \,,
\end{eqnarray}
where $\mat \Sigma = \frac{1}{n} \sum_{i} \hat{\vct x}_{i} \hat{\vct x}_{i}^{\T}$, $\vct \mu = \frac{1}{n} \sum_{i} \hat{\vct x}_{i}$, and $\mat M = \vct x_{c} \vct w^{\T}  + \left(f(\vct x_{c})-y_{c} \right)\mathbb I$.
The term $\vct v$ yields zero for LASSO, the identity matrix $\mathbb I$ for ridge, and $(1-\rho) \mathbb I$ for the elastic net.

The derivatives $\diff{\vct w}{\vct x_{c}}$ and $\diff{b}{\vct x_{c}}$ can be finally obtained by solving the linear system given by Eq.~\eqref{eq:lin-sys}, and then substituted into Eq.~\eqref{eq:gradient-xc} to compute the final gradient.

\begin{algorithm}[tb]
  \caption{Poisoning Embedded Feature Selection}
  \label{alg:poisoning}
  \textbf{Input:} $\hat{\set D}$, the (surrogate) training data; 
  $\{\vct x_{c}^{(0)}, y_{c} \}_{c=1}^{q}$, the $q$ initial attack points with (given) labels; $\beta \in (0,1)$; and $\sigma,\varepsilon$, two small positive constants.\\
  \textbf{Output:} $\{ \vct x_{c} \}_{c=1}^{q}$, the final attack points.
  
  \begin{algorithmic}[1]
    \STATE{$p \gets 0$}
    \REPEAT
        \FOR{$c=1,\ldots,q$}
        \STATE{ $\{ \vct w, b \} \leftarrow $ learn the classifier on $\hat{\set D} \cup \{\vct x_{c}^{(p)}\}_{c=1}^{q}$.}
     \STATE{Compute $\nabla \set W = \diff{\set W (\vct x_{c}^{(p)})}{\vct x_{c}}$  according to Eq.~\eqref{eq:gradient-xc}.}
     \STATE{Set $\vct d = \Pi_{\set B} \left( x_{c}^{(p)} + \nabla \set W \right) - x_{c}^{(p)}$ and $k \gets 0$.}
     \REPEAT[line search to set the gradient step $\eta$]
     \STATE{Set $\eta \gets \beta^{k}$ and $k \gets k+1$}
      \STATE{$\vct x_c^{(p+1)} \gets \vct x_c^{(p)} + \eta \vct d$}
      \UNTIL{$\set W (\vct x_c^{(p+1)}) \leq \set W (\vct x_c^{(p)}) - \sigma \eta \| \vct d \|^{2}  $}
      \ENDFOR
      \STATE{$p \gets p + 1$}
      \UNTIL{$ | \set W (\{\vct x_c^{(p)}\}_{c=1}^{q}) - \set W( \{\vct x_c^{(p-1)}\}_{c=1}^{q})  | < \varepsilon$}
      \STATE{\textbf{return:} $\{\vct x_{c} \}_{c=1}^{q} = \{\vct x^{(p)}_{c} \}_{c=1}^{q}$}
  \end{algorithmic}
\end{algorithm}

\begin{figure}[t]
\centering
\includegraphics[width=0.47\textwidth]{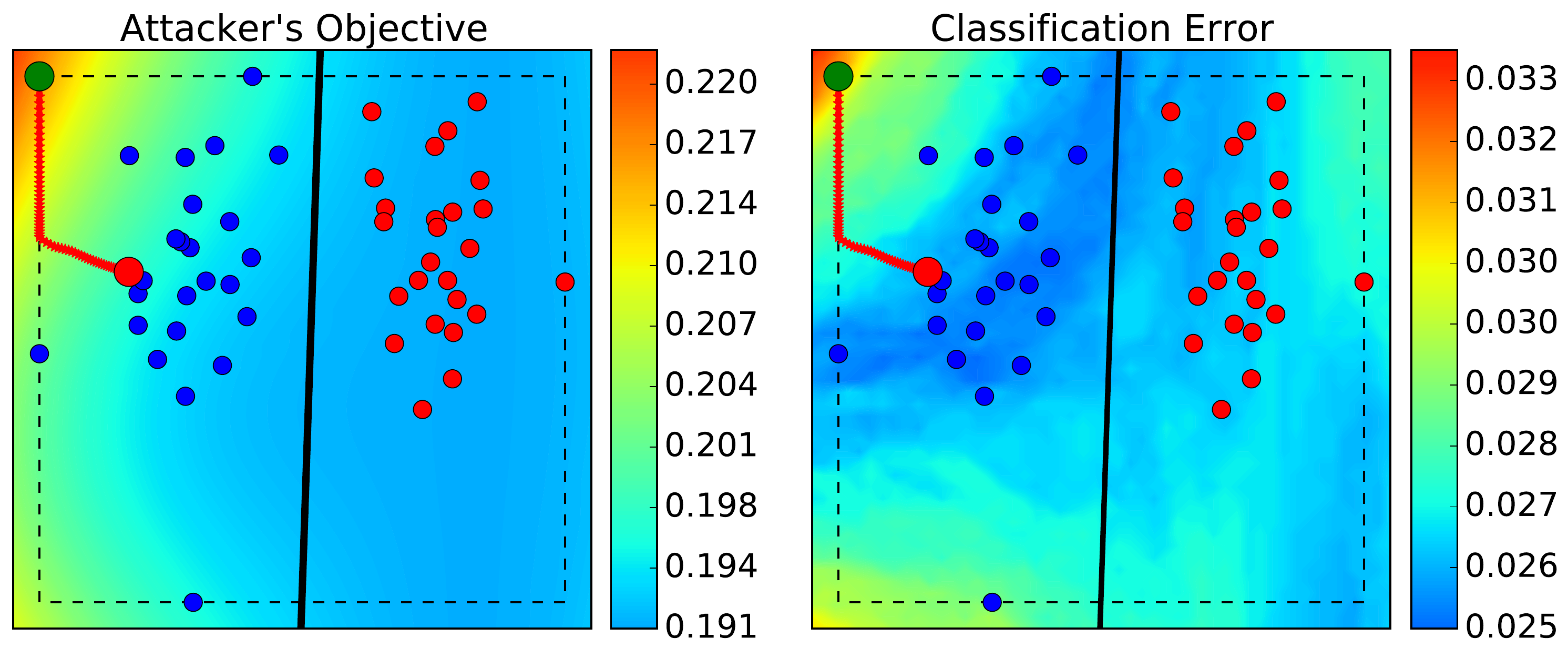}
\vspace{-8pt}
\caption{Poisoning LASSO. Red and blue points are the positive ($y=+1$) and negative ($y=-1$) training data $\set D$. The decision boundary at $f(\vct x) = 0$ (for $\lambda = 0.01$, in the absence of attack) is shown as a solid black line. The solid red line highlights the path followed by the attack point $\vct x_{c}$ (\ie, the magnified red point) towards a maximum of $\set W(\vct x_{c})$ (shown in colors in the \emph{left} plot), which also corresponds to a maximum of the classification error (\emph{right} plot).
A box constraint is also considered (dashed black square) to bound the feasible domain (\ie, the attack space).} 
 \vspace{-5pt}
\label{fig:2d-example}
\end{figure}

\paragraph{Poisoning Feature Selection Algorithm.} The complete poisoning attack algorithm is given as Algorithm~\ref{alg:poisoning}, and an exemplary run on a bi-dimensional dataset is reported in Fig.~\ref{fig:2d-example}. To optimize the attack with respect to multiple attack points, we choose to iteratively adjust one attack point at a time, while updating the current solution $\vct w,b$ at each step (this can be efficiently done using the previous solution $\vct w,b$ as a warm start). This gives the attack much more flexibility than a greedy strategy where points are added one at a time, and never modified after insertion.
We also introduce a projection operator $\Pi_{\set B}(\vct x)$ to project $\vct x$ onto the feasible domain $\set B$; \eg, if features are normalized in $[0,1]$, one may consider $\set B$ as the corresponding box-constrained domain. This enables us to define a feasible descent direction $\vct d$ within the given domain $\set B$, and perform a simple line search to set the gradient step size $\eta$.

\textbf{Descent in Discrete Spaces.} If feature values are discrete, it is not possible to follow the gradient-descent direction exactly, as it may map the given sample to a set of non-admissible feature values. It can be however exploited as a search heuristic. Starting from the current sample, one may generate a set of candidate neighbors by perturbing only those features of the current sample which correspond to the maximum absolute values of the gradient, one at a time, in the correct direction. Eventually, one should update the current sample to the neighbor that attained the maximum value of $\set W$, and iterate until convergence.

\section{Experiments} \label{sect:exp}

In this section, we consider an application example involving the detection of malware in PDF files, \ie, one among the most recent and relevant threats in computer security~\cite{ibm-xforce}.
The underlying reason is that PDF files are excellent carriers for malicious code, due to the flexibility of their logical structure, which allows embedding of several kinds of resources, including \texttt{Flash}, \texttt{JavaScript} and even executable code. 
Resources are simply embedded by specifying their type with \emph{keywords}, and inserting the corresponding content in \emph{data streams}.
For instance, an embedded resource in a PDF may look like:
\vspace{-5pt}
\begin{alltt}
13 0 obj << \textbf{/Kids} [ 1 0 R 11 0 R ]
\textbf{/Type /Page} ... >> end obj
\end{alltt}
\vspace{-5pt}
where keywords are highlighted in bold face.
Recent work has promoted the use of machine learning to detect malicious PDF files (apart from legitimate PDFs), based on the analysis of their logical structure and, in particular, of the present keywords rather than the content of data streams~\cite{smutz12,maiorca12-mldm,maiorca13-asiaccs,srndic13-ndss}. 

\begin{figure*}[htbp]
\centering
\includegraphics[width=0.98\textwidth]{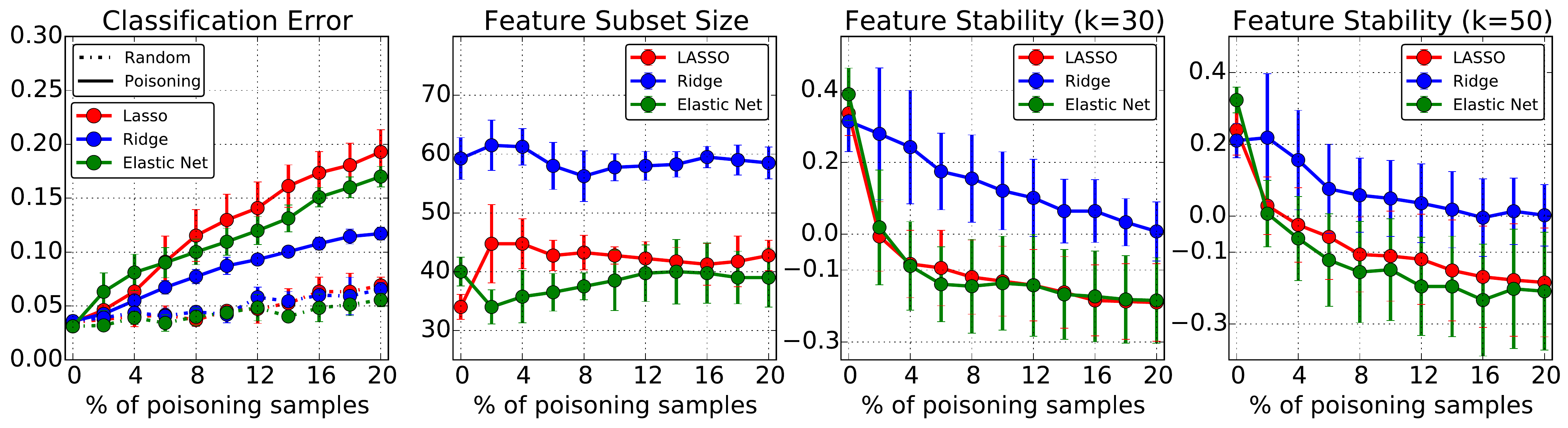}
\includegraphics[width=0.98\textwidth]{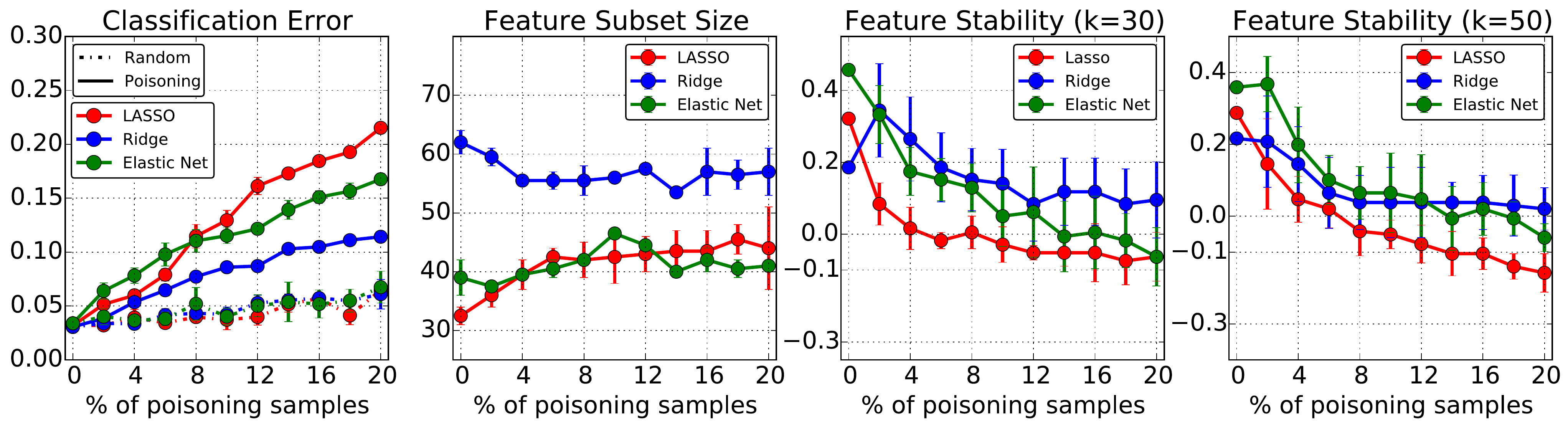}
\vspace{-8pt}
\caption{Results on PDF malware detection, for PK (\emph{top row}) and LK (\emph{bottom row}) poisoning attacks against LASSO, ridge, and elastic net, in terms of classification error (\emph{first column}), number of automatically selected features (\emph{second column}), and stability of the top $k=30$ (\emph{third column}) and $k=50$ (\emph{fourth column}) selected features, against an increasing percentage of injected poisoning samples.
For comparison, we also report the classification error attained by all methods against random label-flip attacks (\emph{first column}).
All the reported values are averaged over five independent runs, and error bars correspond to their standard deviation.}
\label{fig:exp-pdf}
\vspace{-5pt}
\end{figure*}

\textbf{Experimental setup.} In our experiments, we exploit the feature representation proposed by \citet{maiorca12-mldm}, where each feature simply denotes the number of occurrences of a given keyword in the PDF file.
We collected 5993 recent malware samples from the \emph{Contagio} dataset,\footnote{\url{http://contagiodump.blogspot.it}} and 5951 benign samples from the web.
As a preliminary step, following the procedure described by \citet{maiorca12-mldm}, we extracted keywords from the first 1,000 samples in chronological order. The resulting 114 keywords were used as our initial feature set $\set X$.
We then randomly sampled five pairs of training and test sets from the remaining data, respectively consisting of 300 and 5,000 samples, to average
the final results. To simulate LK attacks (Sect.~\ref{sect:attacker-knowledge}), we also sampled an additional set of five training sets (to serve as $\hat{\set D}$) consisting of 300 samples each.
We normalized each feature between 0 and 1 by bounding the maximum keyword count to $20$, and dividing each feature value by the same value. This value was selected to restrict the attacker's capability of manipulating data to a large extent, without affecting generalization accuracy in the absence of attack.\footnote{If no bound on the keyword count is set, an attacker may add an unconstrained number of keywords and arbitrarily influence the training process.}
We evaluate the impact of poisoning against LASSO, ridge and elastic net. 
We first set $\rho=0.5$ for the elastic net, and then optimized the regularization parameter $\lambda$ for all methods by retaining the best value over the entire regularization path~\cite{friedman10,scikit-learn}.
We evaluate our results by reporting the classification error as a function of the percentage of injected poisoning samples, which was increased from 0\% to 20\% (where 20\% corresponds to adding 75 poisoning samples to the initial data).
Furthermore, to understand how feature selection and ranking are affected by the attack, we also evaluate the consistency index originally defined by \citet{kuncheva07-iasted} to evaluate the stability of feature selection under random perturbations of the training data.

\textbf{Kuncheva's Stability Index.} Given two feature subsets $A,B \subseteq \set X$, with $\| A \| = \| B \| = k$, $r = \| A \cap B \|$, and $0 < k < \| \set X \| = d$, it is defined as:
\begin{equation} \label{eq:kuncheva}
I_{C}(A,B) = \frac{rd-k^{2}}{k(d-k)} \in [-1,+1]\, ,
\end{equation}
where positive values indicate similar sets, zero is equivalent to random selections, and negative values indicate strong anti-correlation between the feature subsets.
The underlying idea of this stability index is to normalize the number of common features in the two sets (\ie, the cardinality of their intersection) using a correction for chance that accounts for the average number of common features randomly selected out of $k$ trials.

To evaluate how poisoning affects the feature selection process, we compute this index using for $A$ a feature set selected in the absence of poisoning, and comparing it against a set $B$ selected under attack, at different percentages of poisoning.
To compare subsets of equal size $k$, for each method, we considered the first $k$ features exhibiting the highest absolute weight values.
As suggested by \citet{kuncheva07-iasted}, all the corresponding pairwise combinations of such sets were averaged, to compute the expected index value along with its standard deviation.

\textbf{Experimental results.} Results are reported in Fig.~\ref{fig:exp-pdf}, for both the PK and LK settings. No substantial differences between these settings are highlighted in our results, meaning that the attacker can reliably construct her poisoning attacks even without having access to the training data $\set D$, but only using surrogate data $\hat{\set D}$.
In the absence of attack (\ie, at 0\% poisoning), all methods exhibit reliable performances and a very small classification error.
Poisoning up to $20\%$ of the training data causes the classification error to increase of approximately 10 times, from 2\% to 20\% for LASSO, and slightly less for elastic net and ridge, which therefore exhibit slightly improved robustness properties against this threat.
For comparison, we also considered a \emph{random} attack that generates each attack point by randomly cloning a point in the training data and flipping its label. As one may note from plots in the first column of Fig.~\ref{fig:exp-pdf}, our poisoning strategy is clearly much more effective.

Besides the impact of poisoning on the classification error, the most significant result of our analysis is related to the impact of poisoning on feature selection. In particular, from Fig.~\ref{fig:exp-pdf} (third and fourth column), one can immediately note that the (averaged)  stability index (Eq.~\ref{eq:kuncheva}) quickly decreases to zero (especially for LASSO and the elastic net) even under a very small fraction of poisoning samples. 
This means that, in the presence of few poisoning samples, the feature selection algorithm performs as a \emph{random} feature selector in the absence of attack. In other words, the attacker can almost arbitrarily control feature selection.
Finally, it is worth remarking that, among the considered methods, ridge exhibited higher robustness under attack. We argue that a possible reason besides selecting larger feature subsets (see plots in the second column of Fig.~\ref{fig:exp-pdf}) is that feature weights are more evenly spread among the features, reducing the impact of each training point on the embedded feature selection process. We discuss in detail the importance of selecting larger feature subsets against poisoning attacks in the next section.

\section{Discussion} \label{sect:discussion}

We think that our work gives a two-fold contribution to the state of the art. The first contribution is to the field of adversarial machine learning. We are the first to propose a framework to evaluate the vulnerability of feature selection algorithms and to use it to analyze poisoning attacks against popular embedded feature selection methods, including LASSO, ridge regression, and the elastic net. The second contribution concerns the robustness properties of $\ell_{1}$ regularization. Despite our results are seemingly in contrast with the claimed robustness of $\ell_{1}$ regularization, it is worth remarking that  $\ell_{1}$ regularization is robust against \emph{non-adversarial} data perturbations; in particular, it aims to reduce the variance component of the error by selecting smaller feature subsets, at the expense of a higher bias. Conversely, poisoning attacks induce a systematic \emph{bias} into the training set. This means that an attacker may more easily compromise feature selection algorithms that promote \emph{sparsity} by increasing the bias component of the error. The fact that $\ell_{1}$ regularization may worsen performance under attack is also confirmed by \citet{bo14-nips}, although in the context of evasion attacks. Even if the underlying attack scenario is different, also evasion attacks induce a specific bias in the manipulation of data, and are thus more effective against \emph{sparse} algorithms that exploit smaller feature sets to make decisions.

\section{Conclusions and Future Work}
\label{sect:conclusion}

Due to the massive use of data-driven technologies, the variability and sophistication of cyberthreats and attacks have tremendously increased. In response to this phenomenon, machine learning has been widely applied in security settings. However, these techniques have not been originally designed to cope with intelligent attackers, and are thus vulnerable to well-crafted attacks. 
While attacks against learning and clustering algorithms have been widely analyzed in previous work~\cite{barreno06-asiaccs,huang11,bruckner12,biggio12-icml,biggio13-aisec,biggio14-tkde}, only few attacks targeting feature selection algorithms have been recently considered~\cite{bo14-nips,mei15-aistats,zhang15-tcyb}. 

In this work, we have provided a framework that allows one to model potential attack scenarios against feature selection algorithms in a consistent way, making clear assumptions on the attacker's goal, knowledge and capabilities.
We have exploited this framework to characterize the relevant threat of poisoning attacks against feature selection algorithms, and reported a detailed case study on the vulnerability of popular embedded methods (LASSO, ridge, and elastic net) against these attack.
Our security analysis on a real-world security application involving PDF malware detection has shown that attackers can completely control the selection of reduced feature subsets even by only injecting a small fraction of poisoning training points, especially if \emph{sparsity} is enforced by the feature selection algorithm.

This demands for the engineering of \emph{secure} feature selection algorithms against poisoning attacks.
To this end, one may follow the intuition behind the recently-proposed feature selection algorithms to contrast evasion attacks, \ie, to model interactions between the attacker and the feature selection algorithm~\cite{bo14-nips,wang14-icdm,zhang15-tcyb}. 
Recent work on robust LASSO and robust regression may be another  interesting future direction to implement secure feature selection against poisoning~\cite{nasser11-nips,nguyen13-tit}. 
From a more theoretical perspective, it may be of interest to analyze: ($i$) the impact of poisoning attacks on feature selection in relation to the ratio between the training set size and the dimensionality of the feature set; 
and ($ii$) the impact of poisoning and evasion on the bias-variance decomposition of the mean squared error. These aspects may reveal additional interesting insights also for designing secure feature selection procedures.

\section*{Acknowledgments} This work has been partly supported by the project CRP-59872 funded by Regione Autonoma della Sardegna, L.R. 7/2007, Bando 2012.

\end{document}